\documentclass[10pt,twocolumn]{article}
\usepackage[margin=0.75in]{geometry}
\usepackage[T1]{fontenc}
\usepackage{newtxtext,newtxmath}
\usepackage{amsmath,booktabs,multirow}
\usepackage{graphicx}
\usepackage{url}
\usepackage{enumitem}
\usepackage{caption}
\usepackage{microtype}
\usepackage{placeins}
\usepackage{balance}
\usepackage[colorlinks=true,allcolors=black]{hyperref}
\makeatletter
\renewcommand{\@maketitle}{%
  \newpage
  \null
  \vskip 4pt
  \begingroup
    \centering
    {\fontsize{18}{21}\selectfont\bfseries \@title\par}
    \vskip 10pt
    {\large
      \begin{tabular}[t]{c}\@author\end{tabular}\par}
  \endgroup
  \vskip 12pt
}
\renewcommand{\section}{\@startsection{section}{1}{\z@}%
  {-2ex \@plus .5ex \@minus .2ex}%
  {.7ex \@plus .2ex}{\normalfont\large\bfseries}}
\renewcommand{\subsection}{\@startsection{subsection}{2}{\z@}%
  {-1.5ex \@plus .4ex \@minus .2ex}%
  {.5ex \@plus .2ex}{\normalfont\normalsize\bfseries}}
\makeatother
\title{PUMA: Probability-Constrained Sequential Interaction\\for Forensic Localization}
\author{Zhouli Yu \and Mengyuan Zhou}
\date{}

\begin{document}
\maketitle

\begin{abstract}
Image manipulation localization must jointly exploit semantic structure and noise-sensitive forensic evidence. However, the effects of spatial correspondence, matching quality, and interaction order are often conflated. Existing semantic--forensic fusion methods provide complementary representations, but do not explicitly preserve quality-mismatch information during spatial correspondence or adaptively select the order in which information is propagated across branches. We propose PUMA (Probability-constrained Unbalanced Matching and Aggregation), which combines spatially constrained unbalanced optimal transport with position-wise convex selection between two sequential operators through a shared positional interface. Before normalization, PUMA retains signed transport-mass residuals, constructs row-stochastic cross-branch and within-noise-branch operators, and updates noise features through a gated residual path.

PUMA achieves strong localization results on four test sets, with scores on COVERAGE and NIST16 exceeding the published results of the representative methods listed in our comparison. The experiments support the utility of local matching states and adaptive sequential interaction for cross-source localization and several post-processing conditions.
\end{abstract}

\section{Introduction}
With the rapid development of digital image editing and generative artificial intelligence, the barrier to image manipulation continues to fall. Images altered through splicing, object removal, or generative editing \cite{34,35} are widely circulated on social networks, in news media, and in legal-evidence settings \cite{27}, creating practical risks such as the spread of misinformation and the loss of evidentiary value. Passive forensic techniques that localize manipulation by analyzing an image's content and statistical traces, without relying on an embedded watermark or an original reference image \cite{24}, are therefore important. Pixel-level image manipulation localization requires an algorithm to output a binary mask with the same dimensions as the input, labeling manipulated pixels individually \cite{36}. Compared with image-level authenticity classification, this task provides finer-grained decisions, but it is more challenging because traces can be subtle and scenes are highly diverse \cite{37}.

Early passive-forensics research focused on individual types of manipulation traces and gradually evolved toward the joint use of multiple features \cite{28,7}. Noiseprint models characteristic camera-sensor noise to suppress scene-content interference and extract device-discriminative camera fingerprints, providing a stable microscopic cue for manipulation analysis \cite{5}. CAT-Net jointly models the original RGB image and JPEG-compression artifacts, exploiting the complementarity between content structure and compression traces to improve localization \cite{11,1}. TruFor further combines RGB appearance with a learnable noise-sensitive fingerprint and predicts a reliability measure, improving robustness under complex post-processing \cite{8}. More recently, Co-Transformers separately encode semantic and microscopic forensic evidence and coordinate them through cross-branch feature interaction. It is a representative high-performing paradigm among current two-branch manipulation-localization models \cite{22}.

Although two-branch architectures have substantially improved localization by exploiting complementary semantic and forensic features, current cross-branch interaction designs still have limitations. Many mechanisms use Softmax normalization to produce aggregation weights. This preserves only relative allocation among features and removes information about the overall degree of matching between the two feature sets. In forensic analysis, however, semantic--forensic mismatch itself can be a discriminative cue \cite{29}; normalization discards this potentially valuable information. Conventional attention and balanced optimal transport also assume mass conservation and force all information to be transported. This assumption does not fit manipulated regions, where the two feature types may naturally disagree. Forced transfer of irrelevant information can instead impair forensic representations. In addition, existing methods commonly fix the execution order of within-branch aggregation and cross-branch transport, preventing adaptation to location-dependent differences such as those between normal regions and manipulation boundaries.

To address these issues, we propose Probability-constrained Unbalanced Matching and Aggregation (PUMA). First, a spatially constrained unbalanced optimal transport module replaces conventional balanced matching within a local neighborhood. By relaxing strict mass conservation, it allows the amount of feature transport to vary with the degree of bilateral matching. It also explicitly extracts mass residuals before normalization, turning semantic--forensic mismatch into a usable discriminative cue and recovering information lost by conventional Softmax normalization. Building on this module, we design a probability-constrained sequential interaction block. It computes feature outputs for both ``within-branch aggregation first'' and ``cross-branch transport first'' propagation orders. Using mass residuals as control signals, a position-wise gate adaptively mixes the two outputs instead of imposing a manually fixed order. The interaction remains row-stochastic, and a separate residual-strength gate provides a fallback that prevents mismatched information from interfering with forensic features.

\section{Related Work}
\subsection{Single-source forensic localization}
Early image manipulation localization methods built discriminative representations around specific forensic traces, analyzing local image statistics and anomalous patterns to identify potential manipulated regions \cite{28,9}. Constrained convolution highlights traces left by manipulation by restricting the parameters of the first convolutional layer \cite{49}. Noiseprint models characteristic camera-sensor noise to extract device-discriminative fingerprints and reduce scene-content interference in forensic representations \cite{5}. ManTra-Net localizes multiple types of manipulation by analyzing local anomalies \cite{20,23}. BusterNet targets copy-move manipulation and predicts both source and target regions \cite{50}; DiffForensics uses diffusion priors to learn mesoscopic cues between macro- and microscopic evidence \cite{48}. These methods provide useful evidence from different forensic cues, but their performance remains constrained by trace visibility and image post-processing.

\subsection{Multi-source fusion and interaction}
To exploit complementary evidence, later work moved from single-source representations toward multi-source evidence fusion \cite{16,25}. CAT-Net jointly models RGB information and JPEG-compression-domain features \cite{11}; TruFor combines RGB appearance with noise-sensitive fingerprints and can predict localization reliability \cite{8}. MVSS-Net combines noise and boundary evidence \cite{3,38}. ObjectFormer combines RGB and high-frequency features and uses object prototypes to model regional consistency \cite{47}. Co-Transformers coordinates semantic and microscopic forensic information through dual-branch encoding and cross-branch interaction \cite{22,30}. IML-ViT investigates high-resolution, multi-scale features and edge supervision \cite{45,14}, while Mesorch fuses Transformer and CNN representations in parallel \cite{46}. These studies show that effective multi-source evidence use depends on an appropriate interaction mechanism. PUMA focuses more specifically on two central issues: local transport quality and the order of sequential propagation.

\subsection{Attention and optimal-transport matching}
Attention mechanisms aggregate information through normalized weights over positions or features \cite{21,40} and are widely used for relational modeling and information exchange \cite{41}. Transformers use attention to establish long-range dependencies \cite{17,33}; non-local networks model long-range dependencies through pairwise similarities \cite{19,39}; and graph attention networks perform weighted aggregation over local neighborhoods \cite{18,31}. Unlike similarity-based weighted aggregation, optimal transport represents correspondences between two sets of features through a transport plan. Entropy regularization substantially improves the efficiency of optimal-transport solvers \cite{6}, while unbalanced optimal transport relaxes marginal constraints to handle unequal mass on the two sides \cite{4}. Sinkformers explore Sinkhorn normalization in attention computation \cite{15}. Forensic Similarity compares image regions using forensic traces and emphasizes that forensic correspondence is not equivalent to visual-appearance similarity \cite{13}.

Together, these studies offer different approaches to multi-source fusion through feature weighting, correspondence modeling, and multi-path interaction. Unlike methods that focus on general feature aggregation \cite{42} or correspondence modeling, we investigate the relationship between local matching state and sequential interaction order. At each position, PUMA computes both within-branch-aggregation-first and cross-branch-transport-first outcomes, then uses local matching information to adapt their relative contributions. Spatially constrained unbalanced matching preserves local mass information, while probability-constrained sequential interaction adaptively selects the propagation order.

\section{Method}
This section describes PUMA. Given an input image, the model performs pixel-level manipulation localization. Following Co-Transformers, a two-branch backbone extracts semantic and noise-forensic features. PUMA is a plug-in interaction module inserted at the pre-fusion interface. On a shared spatial grid, it performs feature alignment, spatially constrained unbalanced matching, and adaptive sequential interaction. The two branch features are first projected into a common matching space. Unbalanced optimal transport then establishes cross-branch correspondences and extracts mass-residual cues. Finally, those residuals control the position-wise mixture of two propagation orders and the interaction strength, and the noise features are updated residually. The updated noise features and the original semantic features are passed to the baseline model's existing fusion and localization head. The backbone and prediction pathway remain unchanged \cite{22}.

\begin{figure*}[!t]
\centering
\includegraphics[width=\textwidth]{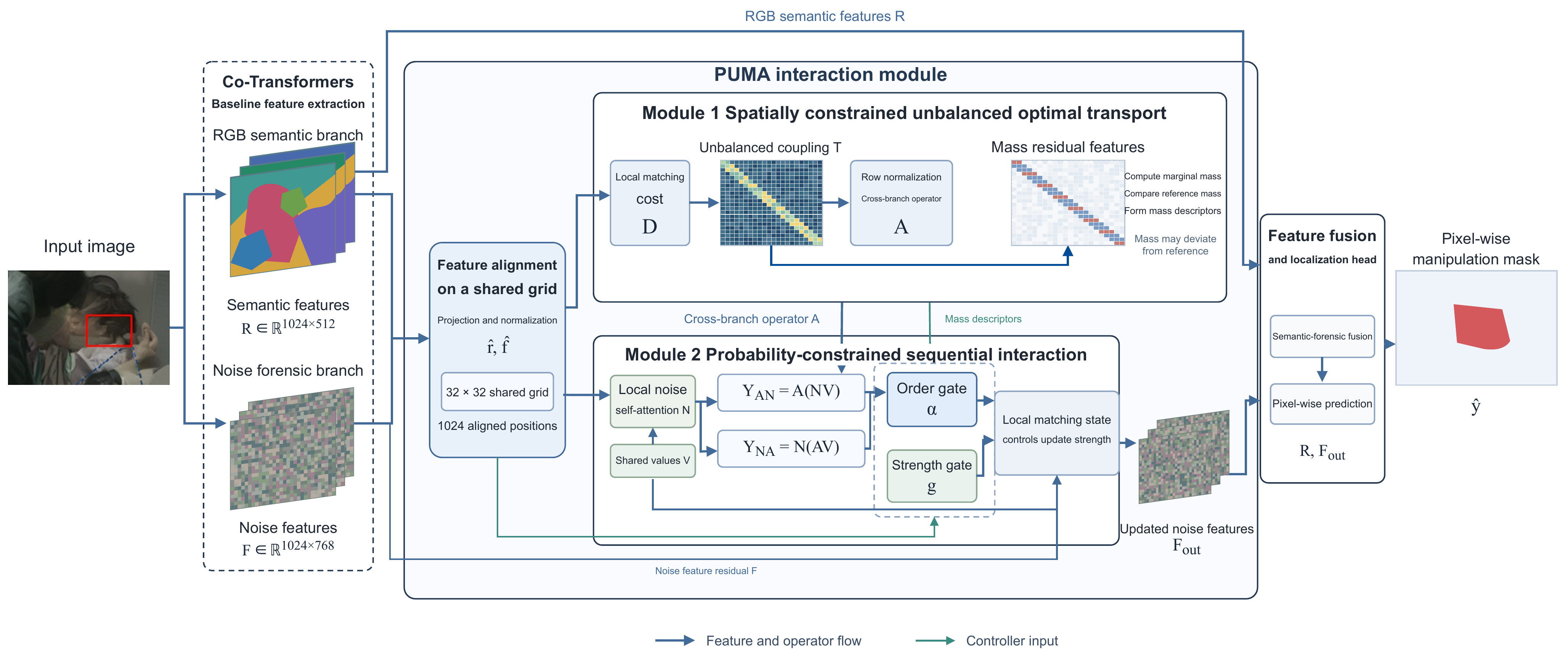}
\caption{Overview of PUMA. Spatially constrained unbalanced matching preserves local mass states, which guide sequential interaction and residual updates of noise features.}
\label{fig:framework}
\end{figure*}

\subsection{Local unbalanced matching and mass residuals}
To address the loss of overall bilateral matching information caused by row normalization and the uninformative forensic transfer that can result from strict mass conservation, this module performs spatially constrained unbalanced optimal-transport matching. Unbalanced transport relaxes mass conservation within a local neighborhood, allowing transported feature mass to vary with bilateral match quality. Before normalization, the module explicitly extracts and retains the mass-residual state. It outputs both relative aggregation weights for cross-branch message passing and match-quality cues for subsequent gated interaction.

At the pre-fusion interface, the RGB feature $R$ and noise feature $F$ have 512 and 768 channels, respectively, and both lie on a $32\times32$ grid. Point-wise matching cannot exploit neighboring responses, whereas global matching can introduce similar textures from distant locations. We therefore set the support radius to 2 and retain only the truncated RGB neighborhood around each noise position, limiting the number of candidates to at most 25. This permits small local offsets while controlling erroneous correspondences and computational cost.

After layer normalization, bias-free linear projection, and unit-norm normalization, both branches yield 64-dimensional descriptors \cite{32,43}. Let $\bar f_i$ and $\bar r_j$ denote the descriptors at noise position $i$ and RGB position $j$, respectively, and let $\mathcal{S}$ denote the local support. The matching cost is
\begin{equation}
\begin{aligned}
D_{ij}&=1-\bar f_i^\top\bar r_j+
\frac{0.1\lVert\Delta_{ij}\rVert_2^2}{2^2},\\
&\hspace{5em}(i,j)\in\mathcal{S}.
\end{aligned}
\end{equation}
The first term measures feature difference using cosine distance, assigning lower cost to descriptors with similar directions. The second term penalizes spatial displacement, where $\Delta_{ij}$ is the grid-coordinate difference and distance is normalized by the squared support radius. The coefficient 0.1 controls the spatial constraint. The local support excludes distant candidates, while the spatial penalty further favors nearby positions, preventing matching from being dominated by appearance similarity alone.

Let $a_i=b_j=1/1024$ be the reference masses at all positions on the two sides, and let the nonnegative matrix $T$ represent transported mass between position pairs. Entries outside the support are fixed to zero. To let weakly corresponding positions transport less mass rather than enforcing fixed row and column totals, we use the following unbalanced optimal-transport objective:
\begin{equation}
\begin{aligned}
\min_{T\geq 0}\; &\langle D,T\rangle
+\varepsilon\,\mathrm{KL}(T\Vert ab^\top)\\
&+\tau_r\,\mathrm{KL}(T\mathbf{1}\Vert a)\\
&+\tau_c\,\mathrm{KL}(T^\top\mathbf{1}\Vert b).
\end{aligned}
\end{equation}
The transport-cost term favors low-cost position pairs. The coupling regularizer uses $ab^\top$ as a reference and controls the concentration of the assignment. The final two terms softly constrain the row and column marginals using generalized KL divergence, allowing actual mass to deviate from the reference mass so that both matching cost and mass deviation determine the transport plan. We set $\varepsilon=0.1$ and $\tau_r=\tau_c=1$, use 128 log-domain iterations, and retain gradients to learn the matching descriptors.

The row and column sums of $T$, denoted by $m_i$ and $n_j$, are the actual transported masses at noise and RGB positions. Their relative residuals are
\begin{equation}
\rho_i=\frac{m_i}{a_i}-1,\qquad
\kappa_j=\frac{n_j}{b_j}-1.
\end{equation}
Positive and negative residuals indicate excess and deficient mass, respectively. These states preserve total-mass differences that row-wise normalization would remove, providing local matching evidence for subsequent gating.

The controller uses $\log(m_i/a_i)$ and $\log(n_j/b_j)$, each centered by subtracting the mean over its grid. The logarithm describes relative change, and centering emphasizes positional differences; the two states are aligned by their shared-grid indices. If a mass ratio is below $10^{-8}$ or non-finite, its log ratio is set to zero before centering. This state describes deviation in matching mass and is not a manipulation probability.

After retaining the mass state, the coupling matrix is row-normalized into a local aggregation operator:
\begin{equation}
A_{ij}=\frac{T_{ij}}{m_i}.
\end{equation}
If $m_i/a_i$ is below $10^{-8}$ or non-finite, that row uses the identity mapping to avoid division by near-zero mass. The operator $A$ is derived from noise--RGB correspondences and is subsequently applied to noise value features indexed on the shared grid, allowing semantic relationships to guide the spatial organization of forensic information. The module thus outputs relative aggregation weights and pre-normalization mass states for message propagation and gate control, respectively.

\subsection{Sequential propagation and gated updates}
Existing cross-branch interaction often fixes the propagation order, which cannot accommodate location-dependent feature relationships at normal regions and manipulation boundaries. Moreover, coupling propagation order with update strength can introduce irrelevant information into forensic representations in mismatched regions. We address these issues with probability-constrained sequential propagation and gated updating. Two operator-composition orders are constructed in parallel; mass residuals from unbalanced matching control a position-wise mixture of the two orders. A separate strength gate decouples order selection from update magnitude, and the noise feature is updated through a residual path while maintaining row-stochastic probability constraints.

On the shared indices and local support of $A$, the noise features are normalized and independently projected to form 64-dimensional queries and keys and a 128-dimensional value field $V$. Query--key dot products are divided by $\sqrt{64}$ and passed through Softmax over the valid neighborhood to obtain the noise self-aggregation operator $N$. The operator $N$ describes within-forensic relationships, while $A$ provides semantic guidance. Both act on the same value field and define two sequential messages:
\begin{equation}
\begin{aligned}
Y_{AN}&=A(NV),\\
Y_{NA}&=N(AV).
\end{aligned}
\end{equation}
The first path supplements local forensic context before propagating messages according to cross-branch relationships; the second first applies semantic guidance and then reorganizes the messages according to forensic relationships. Both paths reuse the same $A$, $N$, and $V$, so their difference is concentrated in the composition order.

Two independent MLPs receive the two centered mass states and the 64-dimensional descriptors from both branches, for an input dimension of 130. Each MLP has a 64-dimensional GELU hidden layer and a Sigmoid output, producing an order gate $\alpha_i$ and a strength gate $g_i$. The descriptors provide content evidence, while the mass states supply matching differences that are unavailable after normalization. The order gate first constructs a position-wise mixed operator:
\begin{equation}
\begin{aligned}
P={}&\operatorname{diag}(\alpha)AN\\
&+\bigl(I-\operatorname{diag}(\alpha)\bigr)NA.
\end{aligned}
\end{equation}
When the order gate approaches 1, it favors $AN$; when it approaches 0, it favors $NA$. Intermediate values yield a continuous selection. To separate ``which order to use'' from ``how much neighborhood information to introduce,'' the strength gate interpolates between mixed propagation and retaining the original value:
\begin{equation}
\begin{aligned}
G&=\operatorname{diag}(g),\\
Q&=I-G+GP,\qquad Z=QV.
\end{aligned}
\end{equation}
A small strength gate preserves more of the original value, whereas a larger one introduces more neighborhood messages. This prevents the order gate from also having to control update magnitude. Both gates are learned from localization supervision; neither mass-residual sign is assigned a fixed switch condition.

Because $A$ and $N$ are nonnegative and row-stochastic, their compositions $AN$ and $NA$ retain these properties. The order and strength gates lie in the unit interval, so the two convex combinations preserve nonnegative spatial weights whose rows sum to one. Thus, order adjustment does not rely on negative weights or amplify the total weight. This probability constraint applies to spatial mixing.

Finally, only the aggregated increment $Z-V$ is mapped back to the 768-dimensional noise space using a bias-free projection $W_o$, followed by a residual update:
\begin{equation}
F_{\mathrm{out}}=F+W_o(Z-V).
\end{equation}
Writing back only the increment exploits neighborhood information while preserving the original forensic representation. As the strength gate approaches zero, the update approaches the identity mapping. The module updates only the noise feature; the RGB feature remains unchanged. They are then concatenated along the channel dimension and passed to the original prediction head.

\subsection{Joint supervision loss}
The proportion of manipulated pixels varies substantially, so optimizing only by pixel count can bias learning toward the background. Regional prediction also requires boundary structure from the semantic branch. Training therefore uses joint region and boundary Dice supervision: the region loss constrains overlap of the final mask, while the boundary loss constrains the structural representation of the RGB branch. The post-interaction forensic information and boundary cues thus jointly support localization.

The final localization output and boundary-branch output are first bilinearly upsampled to the input resolution and passed through Sigmoid to obtain probability maps $p$ and $p_e$. Their labels are the manipulation mask $y$ and its derived boundary mask $y_e$. Both terms use the same Dice loss:
\begin{equation}
\mathcal{L}_{D}(p,y)=1-
\frac{2\sum_{u\in\Omega}p_u y_u+s}
{\sum_{u\in\Omega}p_u^2+\sum_{u\in\Omega}y_u^2+s}.
\end{equation}
Here, $\Omega$ denotes all pixels in a training batch on the current device, and $p_u$ and $y_u$ are the predicted probability and binary label. The smoothing constant is $s=1$ to stabilize computation in cases such as empty regions. The denominator uses the sum of squared predictions and labels; the loss balances regional recall and false-positive control through overall overlap.

Let $\mathcal{L}_{\mathrm{pix}}=\mathcal{L}_{D}(p,y)$ and $\mathcal{L}_{\mathrm{edge}}=\mathcal{L}_{D}(p_e,y_e)$ denote the region and boundary losses, respectively. The total training objective is
\begin{equation}
\mathcal{L}=0.16\,\mathcal{L}_{\mathrm{pix}}
+0.8\,\mathcal{L}_{\mathrm{edge}}.
\end{equation}
The two weights follow the original supervision setup. The region loss backpropagates through the fusion and prediction pathway to optimize matching descriptors, the order gate, and the strength gate. The boundary loss acts on the RGB boundary branch before interaction. The unbalanced-transport objective is used to solve internal correspondences, while the full network is trained end-to-end using the localization supervision above.

\section{Experiments}
\subsection{Data, training, and evaluation}
Training data follow the Protocol-CAT configuration reported by IMDL-BenCo (Ma et al., 2024b) \cite{26}. This protocol builds a training set from manipulated images collected from multiple sources \cite{2,10} and uses source-balanced sampling to reduce training bias caused by differences in source-set sizes.

PUMA and the control group use the same data split, initialization source, supervision, and effective batch size, with the random seed fixed to 42. Both configurations are trained for 150 epochs, and final results are computed from the epoch-150 checkpoint. We use AdamW with an initial learning rate of $1\times10^{-4}$ and weight decay of 0.05. The learning rate is linearly warmed up for the first two epochs and then decayed using a cosine schedule, reaching $5\times10^{-7}$ at the end of training. The effective batch size is 64, consisting of two GPUs, a per-GPU batch size of 16, and two gradient-accumulation steps.

At test time, pixel-level F1 uses a fixed threshold of 0.5. For each image, pixels with predicted probability greater than 0.5 are classified as manipulated, and positive-class F1 is computed as $2TP/(2TP+FP+FN)$. If the denominator is zero, meaning both the ground truth and prediction are empty masks, that image's F1 is set to zero. Each dataset's metric is the arithmetic mean of per-image F1 scores, not a single F1 computed after aggregating pixels across the entire dataset. This protocol prevents true negatives from large background regions from dominating the result and is suitable for localization tasks in which manipulated regions occupy a small fraction of an image.

\begin{figure*}[!t]
\centering
\includegraphics[width=0.82\textwidth]{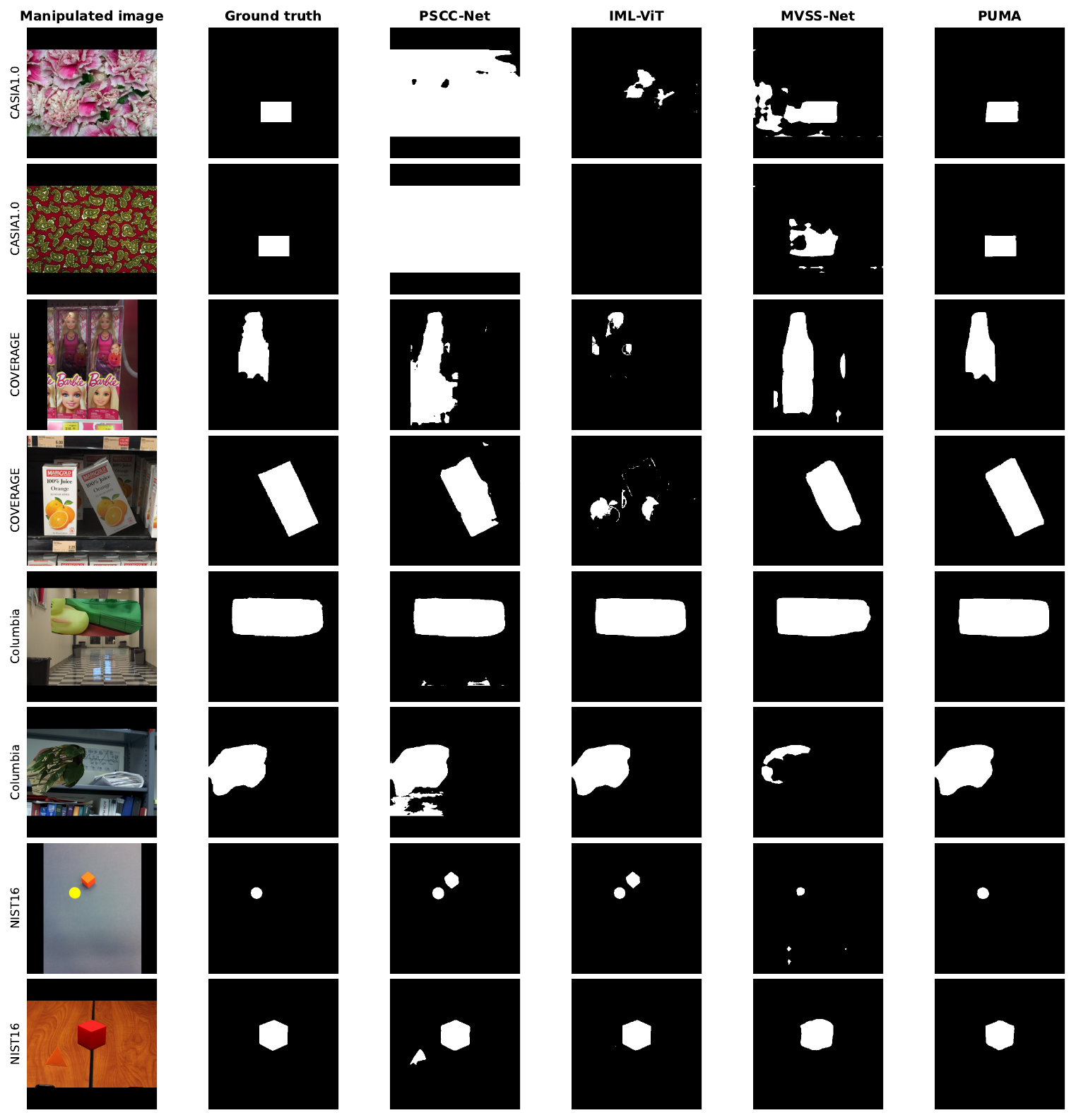}
\caption{Qualitative comparison on four test sets. Each row shows the input image, ground-truth mask, and predictions from PSCC-Net, IML-ViT, MVSS-Net, and PUMA, in that order. Predictions are binarized using a fixed threshold of 0.5.}
\label{fig:qualitative}
\end{figure*}

\subsection{Comparison with existing methods}
PUMA obtains higher pixel-level F1 scores than the corresponding published results listed in Table~\ref{tab:comparison} on all four test sets: CASIA1.0, COVERAGE, Columbia, and NIST16. The gaps are more pronounced on COVERAGE and NIST16. This comparison suggests that local matching and forensic-context interaction are useful under the evaluation protocol adopted here.

\begin{table}[!htb]
\centering
\small
\setlength{\tabcolsep}{3pt}
\renewcommand{\arraystretch}{1.12}
\caption{Published results of representative manipulation-localization methods and results for PUMA \cite{22,44}.}
\label{tab:comparison}
\begin{tabular}{lrrrr}
\toprule
Method & CASIA1.0 & COVERAGE & Columbia & NIST16\\
\midrule
MVSS-Net \cite{3} & 0.5486 & 0.5188 & 0.7322 & 0.2876\\
PSCC-Net \cite{12} & 0.5472 & 0.3791 & 0.8640 & 0.3681\\
IML-ViT \cite{45} & 0.7392 & 0.5364 & 0.9334 & 0.1100\\
CAT-Net \cite{11} & 0.8140 & 0.4272 & 0.9151 & 0.3787\\
TruFor \cite{8} & 0.8176 & 0.4573 & 0.8845 & 0.3480\\
Mesorch \cite{46} & 0.8398 & 0.5862 & 0.8903 & 0.3921\\
\textbf{PUMA (ours)} & \textbf{0.8600} & \textbf{0.6791} & \textbf{0.9358} & \textbf{0.5543}\\
\bottomrule
\end{tabular}
\end{table}

Figure~\ref{fig:qualitative} presents representative examples from the four test sets to compare how methods handle the shape of manipulated regions and false positives in the background. Across several examples, PUMA recovers target boundaries more completely and reduces scattered responses outside the targets compared with the baselines. These observations are consistent with the cross-dataset quantitative results in Table~\ref{tab:comparison}, suggesting that local matching and sequential feature interaction help organize fine-grained forensic information.

\begin{figure*}[!t]
\centering
\includegraphics[width=\textwidth]{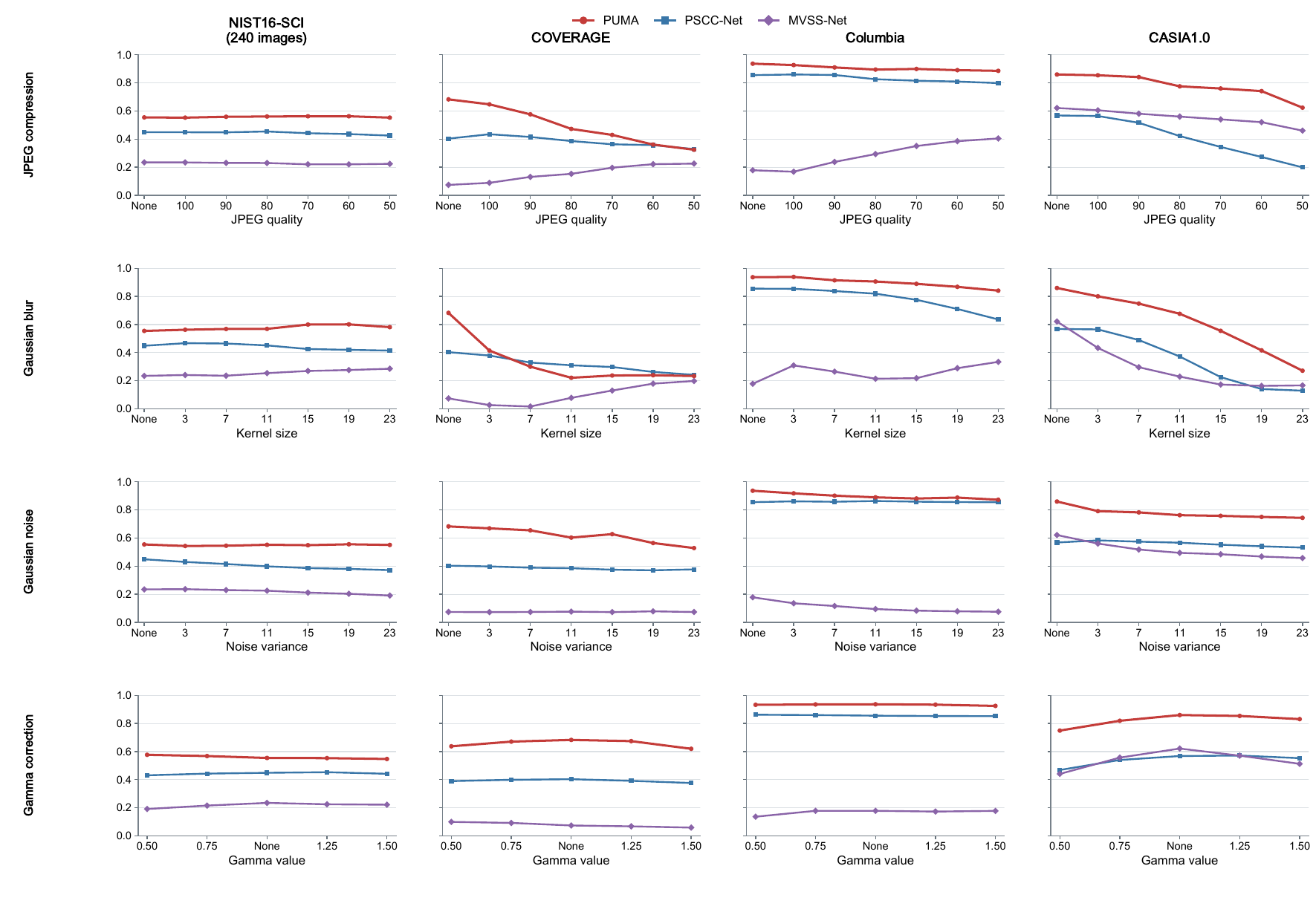}
\caption{Localization robustness comparison among PUMA, PSCC-Net, and MVSS-Net across four test sets and four image perturbations. Rows denote perturbation types and columns denote test sets. The vertical axis is the arithmetic mean of per-image pixel-level F1 at a fixed threshold of 0.5; ``None'' denotes no perturbation.}
\label{fig:robustness}
\end{figure*}

\subsection{Robustness to post-processing}
The evaluation covers four test sets and four image perturbations: JPEG compression, Gaussian blur, Gaussian noise, and Gamma correction. Methods are evaluated on paired test images with the same perturbation types and strengths. Each test image is decoded once, and a shared perturbed version is generated using the specified transformation. Perturbations are applied to the original RGB image before each model's own resizing and normalization pipeline. Gaussian noise is generated with a fixed seed determined by image identifier, perturbation type, and strength, ensuring reproducible paired noisy inputs across methods. All methods use the fixed threshold of 0.5 to compute per-image pixel-level F1, which is then averaged arithmetically across images. Unperturbed results serve as the reference.

Figure~\ref{fig:robustness} shows that PUMA achieves higher per-image pixel-level F1 than PSCC-Net and MVSS-Net under most perturbation conditions across the four test sets. It also maintains relatively stable localization under multiple degradations in Columbia, under compression and noise changes in NIST16-SCI, and under noise and Gamma correction in CASIA1.0. Its advantage narrows under strong JPEG compression and larger blur kernels on COVERAGE and under strong blur on CASIA1.0. Overall, however, the results indicate that PUMA adapts well to a range of post-processing changes.

This behavior is consistent with the interaction mechanism. Unbalanced matching adjusts transport mass according to local match quality and preserves pre-normalization mass deviations as cues for subsequent gating. The order and strength gates use these cues to select a propagation order at each position and control the amount of neighborhood information written back. Feature interaction can therefore adapt to local forensic evidence. When compression, blur, noise, or intensity mapping changes local traces, the model can still use semantic guidance and forensic matching to select and propagate relevant information, reducing interference from mismatched evidence. This provides a mechanism-level explanation for PUMA's advantage over the two comparison methods under most test conditions.

\begin{figure*}[!t]
\centering
\includegraphics[width=\textwidth]{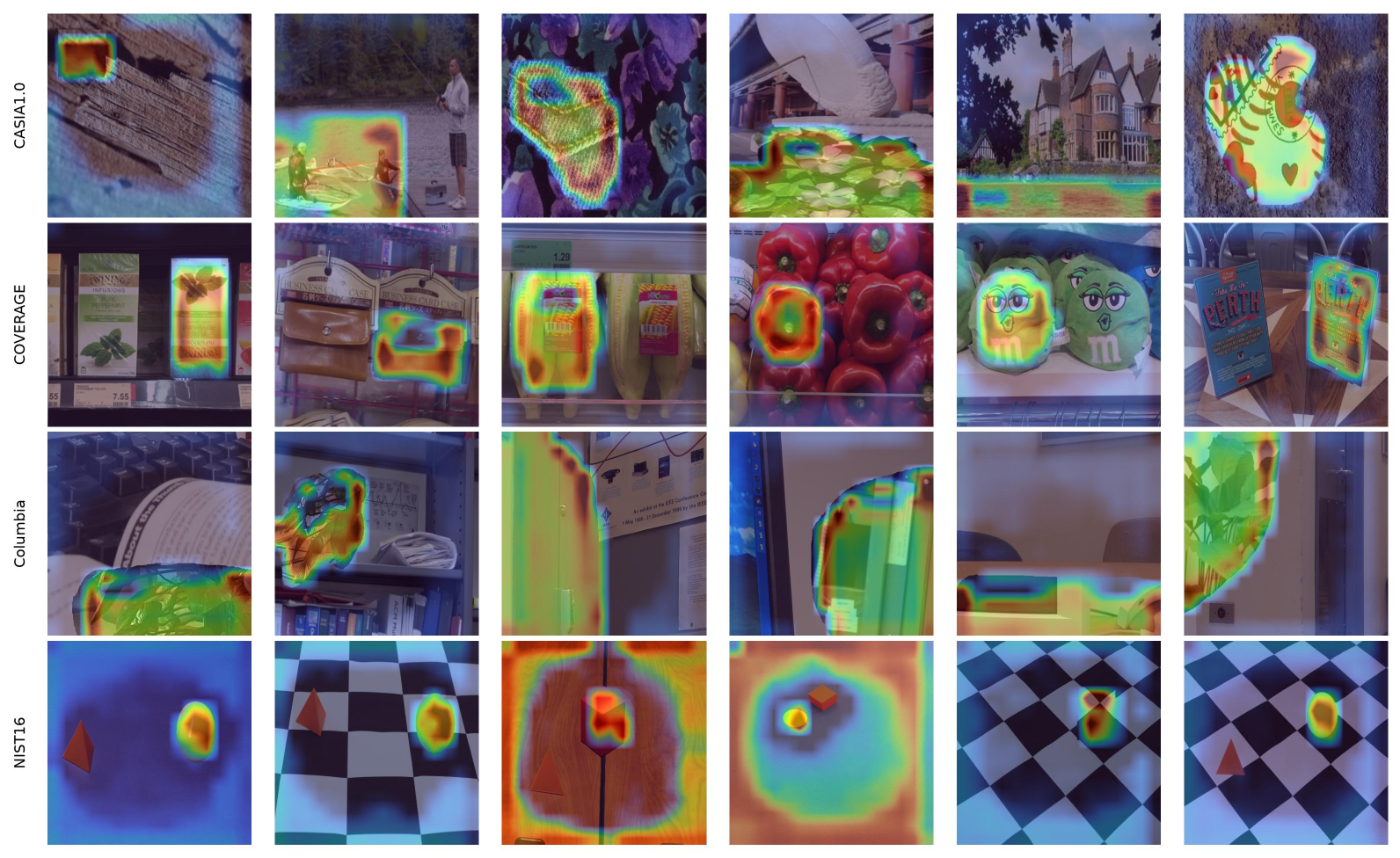}
\caption{Grad-CAM visualizations for PUMA on representative samples from CASIA1.0, COVERAGE, Columbia, and NIST16. Colors indicate the relative response strength of the final-layer features for manipulation-region prediction.}
\label{fig:gradcam}
\end{figure*}

\subsection{Interaction and performance analysis}
Table~\ref{tab:comparison} shows that PUMA exceeds the listed published baseline values on all four test sets, with more pronounced differences on COVERAGE and NIST16. Because these published results were not re-measured in this study using unified weights and data pipelines, the table serves primarily as a reference to existing benchmark values. Structurally, PUMA organizes noise-feature values through semantic correspondences rather than replacing forensic representations directly with RGB features, allowing semantic guidance and forensic evidence to jointly contribute to prediction.

Figure~\ref{fig:gradcam} shows Grad-CAM overlays for representative PUMA examples from the four test sets. Heatmaps are computed from the final backbone features, using the predicted response within the ground-truth manipulated region as the visualization target.

The high-response regions cover manipulated targets or their nearby boundaries in several examples, indicating that the deep features focus on spatial regions relevant to localization. Together with the binary-mask comparisons in Figure~\ref{fig:qualitative}, these visualizations complement the quantitative and qualitative localization results from the perspective of response distribution and provide an intuitive indication that local forensic information contributes to prediction.

The central advantage of unbalanced matching is that it describes local correspondence separately from total matching mass. When overall candidate matching is weak, soft marginal constraints allow transported mass to deviate from the reference mass. The pre-normalization mass state communicates this deviation to the gating module, supplementing the matching information absent from relative aggregation weights. Consequently, the propagation process can use relative neighborhood relationships while adaptively adjusting interaction strength according to differences in local matching state.

The sequential module further decouples propagation order from update strength. The two operator compositions organize local information differently; the order gate adjusts their mixture position by position, while the strength gate controls how much neighborhood increment is written back. Convex combinations and the residual path preserve the contribution of the original feature. The stability of Columbia under compression and blur, and the stable compression response on NIST16, are consistent with the design goal of retaining the original representation while adjusting neighborhood propagation as evidence changes.

\section{Conclusion}
PUMA addresses the interaction between semantic and forensic features in image manipulation localization by integrating spatially constrained unbalanced matching, preservation of mass states, and probability-constrained sequential propagation within a unified local interface. The matching module describes both neighborhood correspondences and deviations in transported mass. The sequential module then adjusts the two propagation orders and their update strength position by position. It enhances the noise representation through a residual update while preserving the row-stochastic property of spatial aggregation weights. The design retains the original two-branch backbone and prediction pathway and provides an effective way to organize forensic information using local matching states.

Across multiple public manipulation-localization benchmarks and common image post-processing perturbations, PUMA achieves competitive localization accuracy under several test conditions, with gains that vary by data distribution and perturbation type. Under clean-image and JPEG-compression conditions, the retrained control performs better on average across datasets, whereas PUMA shows a more pronounced advantage under strong smoothing degradations. The experiments indicate that local mass states preserved by unbalanced matching and the sequential gating mechanism can effectively adjust forensic-information propagation under specific image degradations. Different perturbations, however, place different demands on interaction strategies and require targeted evaluation and optimization.

\FloatBarrier
\end{document}